\documentclass[10pt,twocolumn]{article}
\pdfoutput=1
\usepackage{lipsum}
\usepackage[twocolumn,textwidth=19.00cm,columnsep=.81cm,top=0.9in,bottom=0.9in]{geometry}

\usepackage{amsmath,amssymb,color,graphicx}

\usepackage{color}
\usepackage{tabularx}
\usepackage{booktabs}
\usepackage{enumitem}
\usepackage{multirow}
\usepackage{url}
\usepackage[T1]{fontenc}
\usepackage[rightcaption]{sidecap}
\sidecaptionvpos{figure}{c}
\usepackage[pagebackref=true,breaklinks=true,letterpaper=true,colorlinks,bookmarks=false]{hyperref}

\long\def\ignore#1{}

\begin{document}

\title{Matching Handwritten Document Images}
\author{
\begin{tabular}[t]{c@{\extracolsep{8em}}c} 
Praveen Krishnan  & C.V Jawahar  \\ 
        IIIT Hyderabad & IIIT Hyderabad \\
        \texttt{praveen.krishnan@research.iiit.ac.in} & \texttt{jawahar@iiit.ac.in} \\ \\
\end{tabular}
}
\date{}
\maketitle

\begin{abstract}
We address the problem of predicting similarity between a pair of handwritten document
images written by different individuals. This has applications related to matching 
and mining in image collections containing handwritten content.  A similarity score is computed by detecting patterns
of text re-usages between document images irrespective of the minor variations in
word morphology, word ordering, layout and paraphrasing of the content. Our method does not
depend on an accurate segmentation of words and lines. We formulate the document
matching problem as a structured comparison of the word distributions across two
document images. To match two word images, we  propose a convolutional neural network
(\textsc{cnn}) based feature descriptor. Performance of this representation surpasses the 
state-of-the-art on handwritten word spotting. Finally, we demonstrate the applicability 
of our method on a practical problem of matching handwritten assignments.

%\keywords{Word spotting, handwritten images, plagiarism detection, \textsc{cnn} features.}
\end{abstract}

%On a dataset of documents written by different individuals, we report encouraging results 
%with a mean normalized discounted cumulative gain of 0.89

\section{Introduction}
\label{sec:intro}
\begin{figure}
\centering
\includegraphics[height=2.7cm]{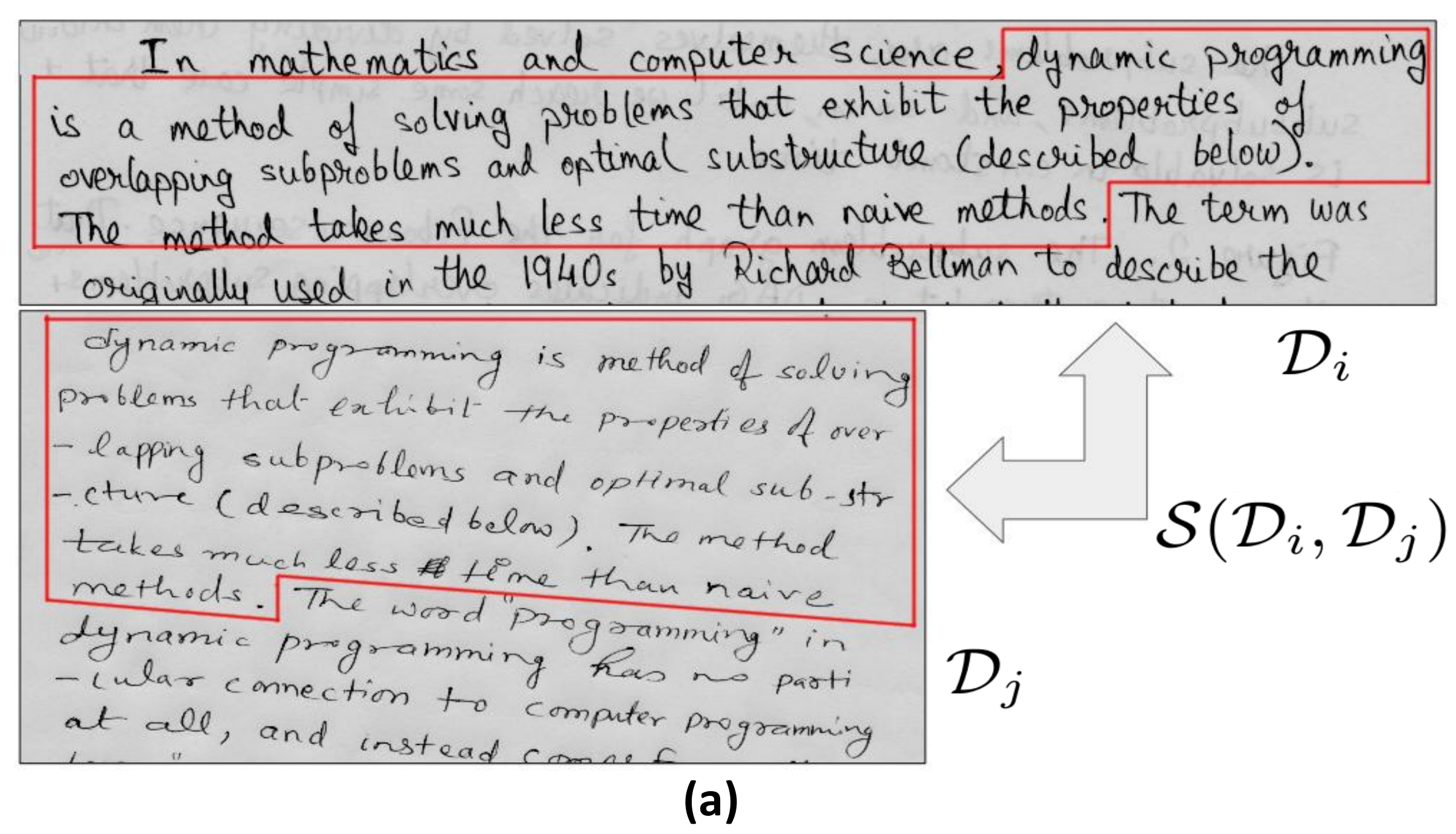}
\includegraphics[height=2.7cm]{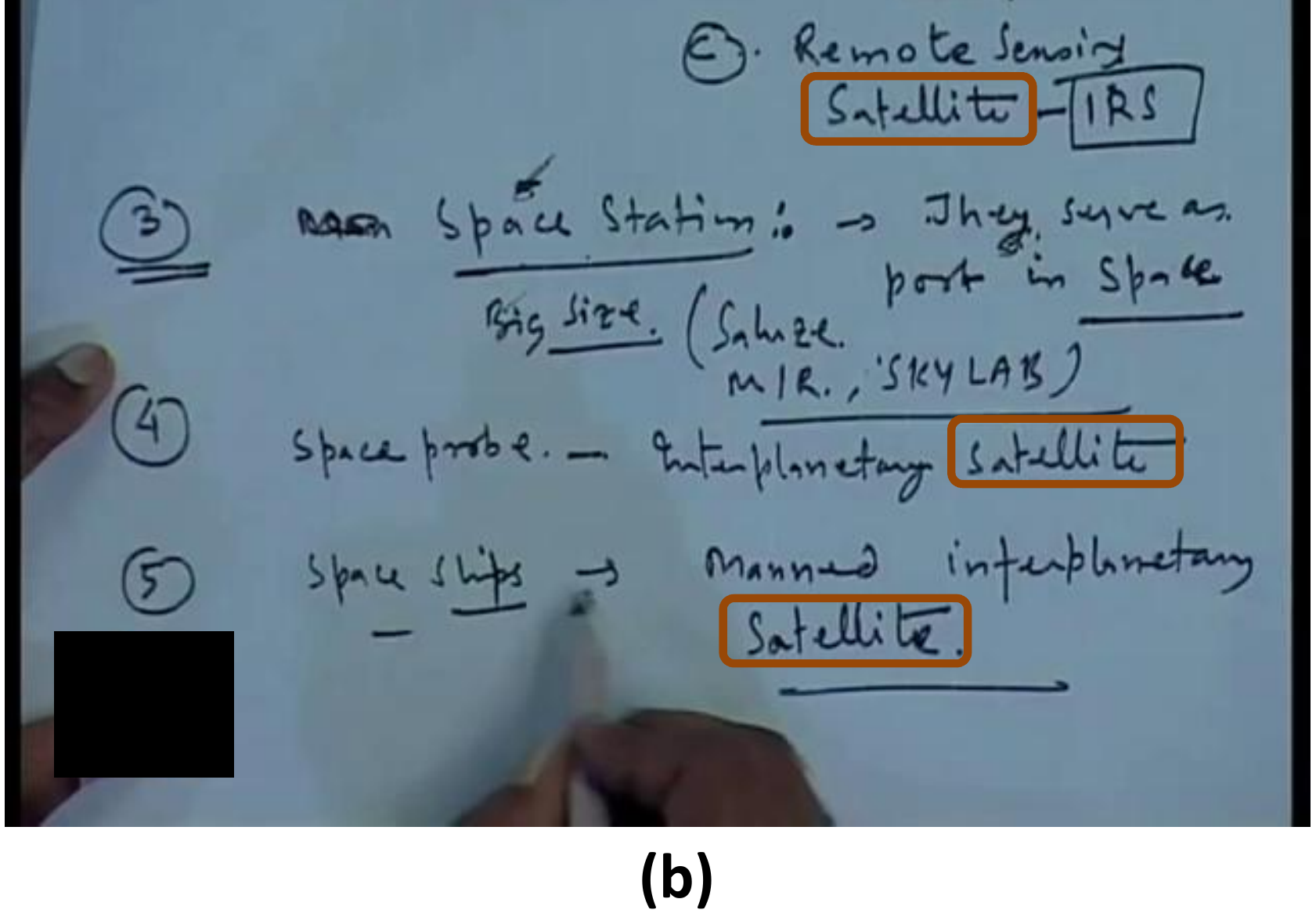}
\caption{(a) Given two document images ${\cal D}_i$ and ${\cal D}_j$, we are interested in computing a similarity score 
${\cal S}({\cal D}_i, {\cal D}_j)$ which is invariant to (i) writers, (ii) word flow across lines, (iii) spatial 
shifts, and (iv) paraphrasing. In this example, the highlighted lines from ${\cal D}_i$ and ${\cal D}_j$ have almost the same 
content but they widely differ in terms of spatial arrangement of words. (b) Query-by-text results on searching with ``satellite''
on an instructional video. The spotted results are highlighted in the frame}
\label{fig:hwmatch}
\end{figure}

Matching two document images has several  applications related to information retrieval like
spotting keywords in historical documents~\cite{fernandez2014sequential},
accessing personal notes~\cite{ManmathaHR96}, camera based interface for querying~\cite{TakedaDAS12},
retrieving from video databases~\cite{MishraAJ13}, automatic scoring of answer sheets~\cite{SrihariCSSSB08}, and
mining and recommending in health care documents~\cite{milewski2004handwriting}.
Since {\sc ocr}s do not reliably work for all types of documents,
one resorts to image based methods for comparing textual content. 
This problem is even more complex when considering unconstrained handwritten documents due 
to the high variations across the writers. Moreover, variable placement of the words across documents makes 
a rigid geometric matching ineffective. In this work,
we design a scheme for matching two handwritten document images. The problem is illustrated in Fig.~\ref{fig:hwmatch}(a). We
validate the effectiveness of our method on an
application, named as measure of document similarity {\sc (mods)}.\footnote{In parallel to measure of 
software similarity {\sc (moss)}~\cite{SchleimerWA03}, which has emerged as the de facto standard across 
the universities to compare
two software solutions from students.} \textsc{mods} compares two handwritten
document images and provides a normalized score as a measure of similarity
between two images.
%\begin{SCfigure}
%\centering
%\caption{Given two document images ${\cal D}_i$ and ${\cal D}_j$, we are interested in computing a similarity score 
%${\cal S}({\cal D}_i, {\cal D}_j)$ which is invariant to (i) writers, (ii) word flow across lines, (iii) spatial 
%shifts, and (iv) paraphrasing. In this example, the highlighted lines from ${\cal D}_i$ and ${\cal D}_j$ have almost the same 
%content but they widely differ in terms of spatial arrangement of words}
%\includegraphics[height=3.5cm]{Images/Motivation.pdf}
%\label{fig:hwmatch}
%\end{SCfigure}

%\noindent \textbf{Text and images.} Understanding text from images...
%\vspace{5cm}
Text is now appreciated as a critical information in understanding natural images~\cite{WangB10,MishraCVPR12,jaderberg2014IJCV}.
Attempts for wordspotting in natural images~\cite{WangB10} have now matured to end-to-end 
frameworks for recognition and retrieval~\cite{WangBB11,JaderbergECCV14,jaderberg2014IJCV}. 
%They use either the energy minimization solutions on a CRF framework on deep learning. 
Natural scene text is often seen as an isolated character image sequence in arbitrary view points or font styles. 
Recognition in this space is now becoming reliable, especially with the recent attempts that use {\sc cnn}s and 
{\sc rnn}s~\cite{jaderberg2014IJCV,SuL14}. However, handwritten text understanding is still lacking in many aspects. 
For example, the best performance on the word spotting (or retrieval) on the hard {\sc iam} data set is an mAP 
of $0.55$~\cite{AlmazanPAMI14}. In this work, we improve this to $0.80$. We achieve this with the 
help of a new data set that now enables the exploitation of deep learnt representations for handwritten data. 

%Beyond the 
%word spotting, we make the following contributions: (i) a novel morphologically invariant word spotting
%is presented. (ii) we demonstrate the application for text in instructional videos.
%(iii) We also demosntrate the use in plagiarism detection in homeworks.

%Word spotting for scene text images is also popular and one of the earliest works in this field came with~\cite{WangB10}.
%More recently in the same context of natural scene text images, deep learnt features have shown potential in matching word images
%~\cite{GoodfellowBIAS13,JaderbergSVZ14,JaderbergECCV14,jaderberg2014IJCV}. As compared to these features
%the proposed \textsc{cnn} descriptor perform better in multi-writer scenarios for \textsc{hw} documents and are more robust to
%translation, skew and distortions.
\vspace{0.2cm}
\noindent \textbf{Word Spotting.} 
%Previous methods in matching documents often limited the scope to matching the word images.
Initial attempts for matching handwritten words were based on \textsc{dtw}~\cite{RathM07} and
\textsc{hmm}~\cite{Fischer12,Rodriguez-SerranoP12}  over variable length feature representations.
Although these models were flexible, they were not really scalable. Many approaches such 
as~\cite{PerronninR09,Rusinol15,AlmazanPR14} demonstrated word spotting using fixed length 
representation based on local features such as \textsc{sift} and \textsc{hog}
along with the bag of words (\textsc{bow}) framework. Most of these works employed better feature
representations such as Fisher vectors~\cite{PerronninR09,AlmazanPR14}, latent semantic
indexing~\cite{Rusinol15}, feature compression~\cite{AlmazanPR14} and techniques such as
query expansion and re-ranking for enhancing the performance. However, the applicability of these methods 
are still limited for multi-writer scenarios. Recently, Almaz{\'{a}}n {\em et al.}~\cite{AlmazanPAMI14} 
proposed a label embedding and attributes learning framework where both word images and text strings 
are embedded into a common subspace with an associated metric to compare both modalities. 
%We also
%demonstrate the utility in retrieving relevant video frames from instructional videos.
%Beyond the
%popular document image collections, we also show the applicability for educational videos containing
%handwritten content.

\begin{figure*}[t]
\centering
\setlength\heavyrulewidth{0.15em}
\setlength\lightrulewidth{0.15em}
\begin{tabularx}{\textwidth}{@{} c | X | l@{}}
\toprule
Query & Top ranked retrieval & Scope\\
\midrule
\fbox{\includegraphics[width=1.9cm,height=0.7cm]{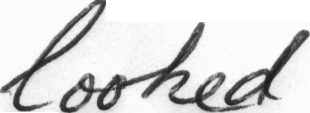}} &
\fbox{\includegraphics[width=1.9cm,height=0.7cm]{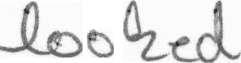}}
\fbox{\includegraphics[width=1.9cm,height=0.7cm]{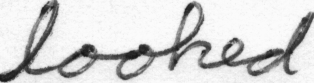}}
\fbox{\includegraphics[width=1.9cm,height=0.7cm]{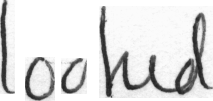}}
\fbox{\includegraphics[width=1.9cm,height=0.7cm]{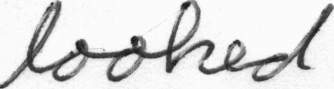}}
\fbox{\includegraphics[width=1.9cm,height=0.7cm]{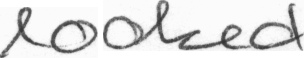}}
\fbox{\includegraphics[width=1.9cm,height=0.7cm]{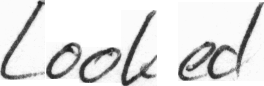}}&
\fbox{\includegraphics[width=2cm,height=0.7cm]{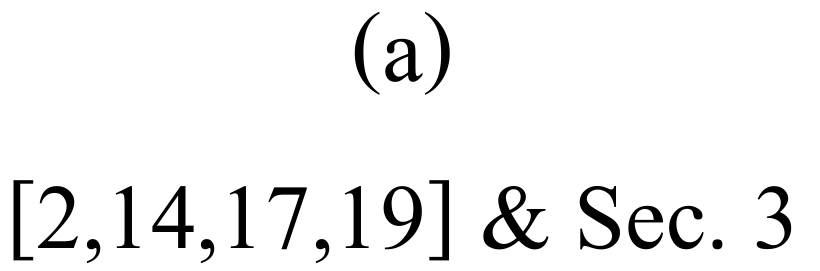}}
%(a)\cite{AlmazanPAMI14,ManmathaHR96,PerronninR09,AlmazanPR14} \& Sec.\ref{sec:hws}
\\
\fbox{\includegraphics[width=1.9cm,height=0.7cm]{Images/VisStem/3.png}} &
\fbox{\includegraphics[width=1.9cm,height=0.7cm]{Images/VisStem/4.png}}
\fbox{\includegraphics[width=1.9cm,height=0.7cm]{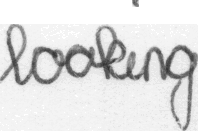}}
\fbox{\includegraphics[width=1.9cm,height=0.7cm]{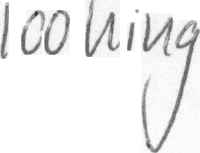}}
%\fbox{\includegraphics[width=1.1cm,height=0.7cm]{Images/VisStem/9.png}}
\fbox{\includegraphics[width=1.9cm,height=0.7cm]{Images/VisStem/10.png}}
\fbox{\includegraphics[width=1.9cm,height=0.7cm]{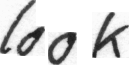}}
\fbox{\includegraphics[width=1.9cm,height=0.7cm]{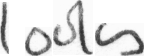}}&
%(b) Our work
\fbox{\includegraphics[width=2cm,height=0.7cm]{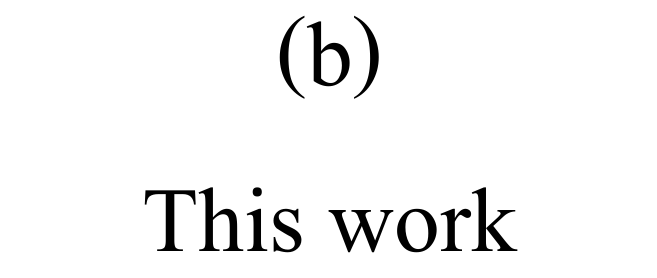}}
\\
\bottomrule
\end{tabularx}
\caption{Word spotting vs. normalized word spotting. (a) shows the conventional word spotting task 
while (b) extends the task to retrieve semantically similar words using a normalized representation. Here
we deal with popular inflectional ending present due to agglutinative property of a language 
which changes the tense, voice of the original words}

\label{fig:normWordSpot}
\end{figure*}

\vspace{0.2cm}
\noindent \textbf{Matching documents.} Matching textual documents is a well studied problem in text processing~\cite{ManningIR08} with
applications in plagiarism detection in electronic documents~\cite{PotthastCELF14}. For softwares, 
{\sc moss}~\cite{SchleimerWA03} provides a solution to compare two programs and is robust against a set of 
alterations e.g., formatting and changes in variable names. However, when the documents are scanned images, 
these methods can not be directly applied. There have been some attempts~\cite{SukthankarMM04,ChumBMVC08} 
to find duplicate and near duplicates in multimedia databases. However, they are not directly applicable 
to documents where the objective is to compare images based on the textual content. For printed documents,
matching based on geometry or organization of a set of keypoints has been successful
~\cite{TakedaICDAR11,GandhiICDAR13,VitaladevuniCPN12}. This works well for duplicate 
as well as cut-and-paste detection in printed documents. However, due to unique set of challenges in handwritten 
documents such as wide variation of word styles, the extraction of 
reliable keypoints with geometric matching is not very successful. Other major challenges include paraphrasing of the 
textual content, non-rigidity of word ordering which leads to word overflows across lines. In our proposed method, 
we uses locality constraints to achieve invariance to such variations. We also extend the word spotting to take care 
of the popular word morphological variations in the image space as shown in Fig.~\ref{fig:normWordSpot}(b). The proposed 
features can associate similarity between word images irrespective of word morphological variations due to changes in 
tense and voice of the sentence construction. In the context of retrieval systems it improves the recall to
search queries and also helps in matching documents in a semantic space.

%We also propose a scheme for matching two handwritten documents with robustness to the natural variations 
%in handwriting. 

\vspace{0.2cm}
\noindent \textbf{Contributions.} In this work, we compute a similarity score by detecting patterns of text re-usages across documents 
written by different individuals irrespective of the minor variations in word forms, word ordering, layout or 
paraphrasing of content. In the process of comparing two document images, we design a module that compares two handwritten 
words using {\sc cnn} features and report a $55\%$ error reduction in word spotting task on the challenging dataset of \textsc{iam} 
and pages from \textsc{gw} collection. We also propose a normalized feature representation for word images which is invariant to different 
inflectional endings or suffixes present in words. 
The advantage of our matching scheme is that it does not require an accurate segmentation of the 
documents. To calibrate the similarity score with that of human perception, we conduct a human experiment where a set of 
individuals are advised to create similar documents with natural variations. Our solution reports a score that match 
the human evaluation with a mean normalized discounted cumulative gain ($nDCG$) of $0.89$. Finally, we demonstrate  
two immediate applications (i) searching handwritten text from instructional videos, and (ii) comparing handwritten 
assignments. Fig.~\ref{fig:hwmatch}(a,b) shows a sample result from these applications.
%The {\sc mods} system captures the handwritten documents using a 
%mobile app which scans the physical pages, and uploads it to a central server where document images are matched.

\section{CNN features for handwritten word images}
\label{sec:wordSpotting}

\begin{figure}[t]
\centering
\includegraphics[height=3cm,width=8cm]{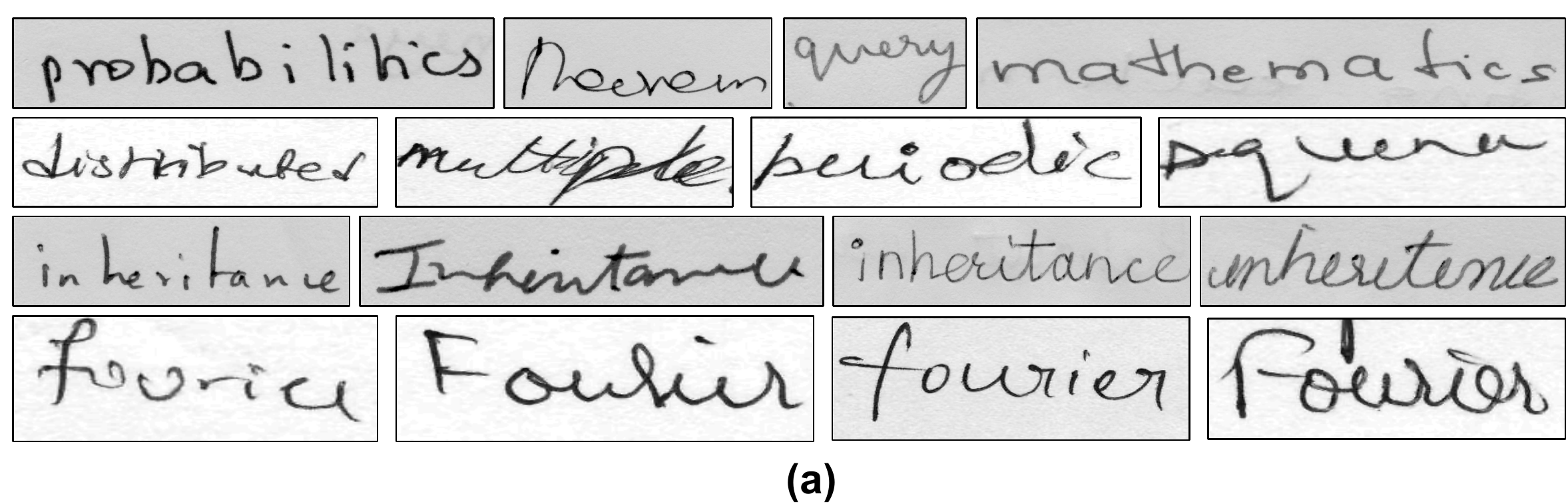}
%\label{fig:sampleWordImages}
\includegraphics[height=3cm,width=8cm]{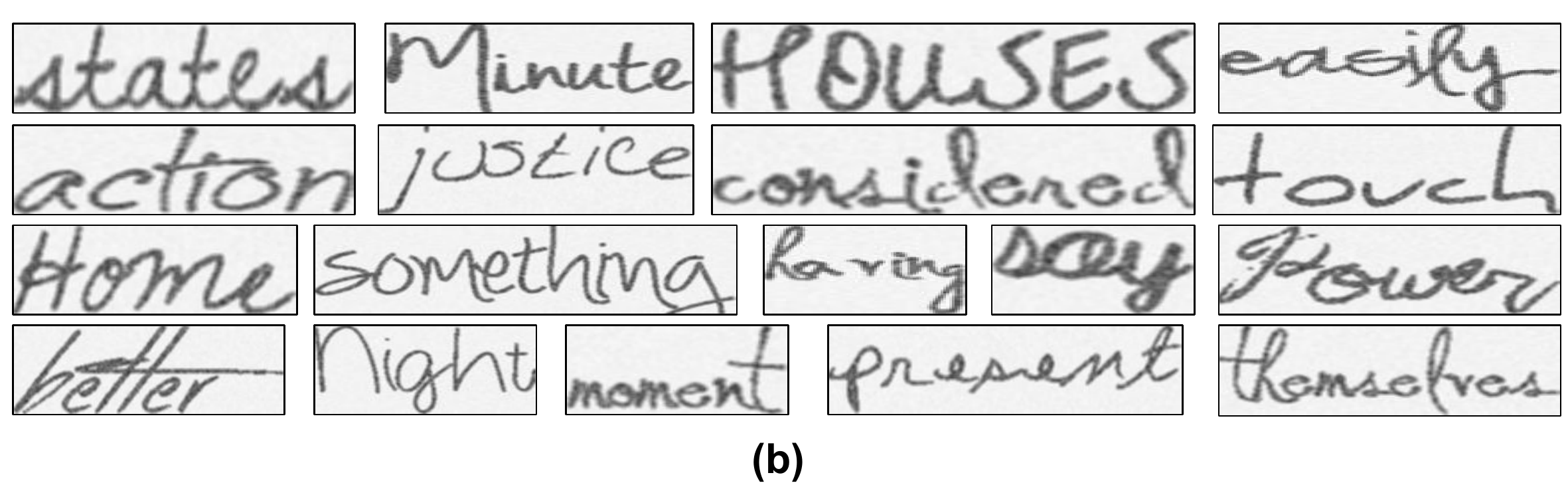}
%\label{fig:sampleSyn}

\caption{(\textbf{a}) The top two rows show the variations in handwritten word images taken from our document collection,  
the bottom two rows demonstrate the challenges such as intra class variability of word images across writers. 
(\textbf{b}) Sample word images from the {\sc hw-synth} dataset created as part of this work to address the lack of training data for 
for learning complex \textsc{cnn} networks}

\label{fig:wordImages}
\end{figure}

The proposed document image matching scheme employs a discriminative representation for
comparing two word images. Such a representation needs to be invariant to (i) both inter and 
intra class variability across writers, (ii) presence of skew, (iii) quality of ink, and (iv) 
quality and resolution of the scanned image. Fig.~\ref{fig:wordImages}(a) demonstrates the 
challenges in matching across writers and documents.
The top two rows show the variations across images in which some are even hard for humans to read 
without enough context of nearby words. The bottom two rows show different instances of same word 
written by different writers, e.g., ``inheritance'' and ``Fourier'' where one can clearly notice the 
variability in shape for each character in the word image. In this work we use convolutional neural 
networks \textsc{(cnn)} motivated by the recent success of deep neural 
networks~\cite{JaderbergSVZ14,Simonyan15,SzegedyLJSRAEVR15,DengDSLL009,KrizhevskySH12} 
and the availability of better learning schemes~\cite{hinton2012improving,IoffeS15}.
Even though \textsc{cnn} architectures such as~\cite{Lecun98,SimardSP03} were among the first to show high
performing classifier for \textsc{mnist} handwritten digits, application of such ideas for
unconstrained continuous handwritten words or documents has not been demonstrated possibly due to the
lack of data, and also the lack of appropriate training schemes. 

\subsection{The HW-SYNTH dataset}
To address the lack of data for training handwritten word images, we build a synthetic handwritten dataset of 
1 million word images. We call this dataset as {\sc hw-synth}. Some of the sample images from this dataset 
are shown in Fig.~\ref{fig:wordImages}(b). Note that these images are very similar to natural handwriting. 
The {\sc hw-synth} dataset is formed out of 750 publicly available handwritten fonts. We use a subset of 
popular Hunspell dictionary and pick a unique set of 10K words for this purpose. For each word, we randomly 
sample 100 fonts and render its corresponding image. 
During this process, we vary the following parameters: (i) kerning level (inter character space), 
(ii) stroke width, and (iii) mean foreground and background pixel distributions. We also perform Gaussian 
filtering to smooth the final rendered image. Moreover, 
we prefer to learn a case insensitive model for each word category, hence we perform three types of rendering,
namely, all letters capitalized, all letters lower and only the first letter in caps.
%\footnote{Code for generation of the dataset
%as well as the trained~\textsc{cnn} models are available at {\tt http:will.be.released.later.}}

\subsection{HWNet architecture and transfer learning}
\label{subsec:hwnet}
The underlying architecture of our \textsc{cnn} model (HWNet) is inspired from~\cite{JaderbergSVZ14}. We use 
a~\textsc{cnn} with five convolutional layers with 64, 128, 256, 512 and 512 square filters with dimensions: 
5, 5, 3, 3 and 3 respectively. The next two layers are fully connected ones with 2048 neurons each. The last 
layer uses a fully connected (FC) layer with dimension equal to number of classes, 10K in our case, and 
is further connected to the softmax layer to compute the class specific probabilities. Rectified linear units 
are used as the non-linear activation units after each weight layer except the last one, and $2 \times 2$ max pooling 
is applied after first, second, and third convolutional layers. We use a stride of one and padding is done 
to preserve the spatial dimensionality. We empirically observed that the recent approach using batch 
normalization~\cite{IoffeS15} for reducing the generalization error, performed better as compared to dropouts. 
The weights are initialized randomly from normal distribution, and during training the learning rate is reduced 
on a log space starting from 0.1. The input to the network is a gray scale word image of fixed size $48\times128$. 
HWNet is trained on the {\sc hw-synth} dataset with 75-15-10\% train-validation-test split using a multinomial logistic 
regression loss function to predict the class labels, and the weights are updated using mini batch gradient descent algorithm with momentum.

\noindent \textbf{Transfer learning.} It is well-known that off-the-shelf 
\textsc{cnn}s~\cite{DonahueJVHZTD14,RazavianASC14} trained for a related task could be adapted or fine-tuned 
to obtain reasonable and even state-of-the-art performance for new tasks. In our case we prefer to perform a transfer learning 
from synthetic domain ({\sc hw-synth}) to real world setting. Here we use popular handwritten labeled corpora such 
as {\sc iam} and {\sc gw} to perform the transfer learning. It is important to keep the learning rates low 
in such setting, else the network quickly unlearns the generic weights learned in the initial layers. In this
work, we extract the features computed from the last FC layer to represent each handwritten word image.
In the supplementary material, we show the advantages of such a network compared to one 
trained from scratch over {\sc hw} corpus. We use MatConvNet toolbox~\cite{vedaldi15matconvnet} 
for training and fine tuning \textsc{cnn} models on top of NVIDIA K40 GPU.

\section{Normalized word spotting}
\label{sec:hws}
Word spotting~\cite{ManmathaHR96,AlmazanPAMI14} has emerged as a popular framework for search and retrieval of text in images. 
It is also considered as a viable alternative to optical character recognition \textsc{(ocr)} in scenarios where the 
underlying document is difficult to segment, e.g., historical manuscripts, handwritten 
documents where the performance of \textsc{ocr} is still limited. Word spotting is typically formulated as a retrieval 
problem where the query is an exemplar image (query-by-example) and the task is to 
retrieve all word images with similar content. It uses holistic word image representation which does not demand character level segmentation.
Some of the popular features include word profiles~\cite{ManmathaHR96}, Fisher vectors~\cite{AlmazanPR14} and the retrieval is performed
using \textsc{knn} search.
%distinguishes it from \textsc{ocr}
%based framework which requires strict character level segmentation. 
Fig.~\ref{fig:normWordSpot}(a) 
shows a word spotting result which retrieves similar word images for the query ``looked''. In this work, our 
interest lies in finding the document similarity between a pair of handwritten documents written by different writers in 
an unconstrained setting. We observe that such a problem can be addressed in a word spotting framework where the task would be
to match similar words between a pair of documents using the proposed \textsc{cnn} features for handwritten word images.

%\vspace{-0.3cm}
%\subsection{Normalized word spotting}
%\label{subsec:normWordSpot}
HWNet provides a generic representation for word spotting by retrieving 
word images with the exact content written. While addressing the larger problem of 
document retrieval, on similar lines of a text based information retrieval pipeline, we relax this constraint 
and prefer to retrieve not just similar or exact words but also their common variations. These variations are 
observed in languages due to morphology. In English, we observe such 
variations in the form of inflectional endings (suffixes) such as ``-s (plural), -ed (past tense), -ess (adjective), -ing 
(continuous form)'' etc. These suffixes are added to the root word, and thereby resulting in a semantically related word. 
A stemmer, such as the Porter stemmer~\cite{porter80} can strip out common suffixes which generates
a normalized representation of words with common roots. We imitate the process of stemming in the visual domain by labeling
the training data in terms of root words given by the Porter stemmer, and use the HWNet architecture to learn
a normalized representation which captures the visual representation of word images along with the invariance to its
inflectional endings. We argue that such a network learns to give less weights to popular word suffixes and gives
a normalized representation which is better suited for document image retrieval tasks. Fig.~\ref{fig:normWordSpot}(b) shows the normalized 
word spotting results obtained using the proposed features that includes both ``similar'' and ``semantically-similar'' 
results, e.g., ``look'',``looks'', ``looking'' and ``looked''.

\section{Measure of document similarity}
\label{sec:docMatch}
Matching printed documents for retrieving the original documents and 
detecting cut-and-paste for finding plagiarism were attempted in the 
past by computing interest points in word images and their corresponding matches~\cite{TakedaICDAR11,GandhiICDAR13}. 
However, handwritten 
documents have large intra class variability to reliably detect interest 
points. In addition, the problem of word-overflow in which words from the 
right end of the document overflow and appear on the left 
end of the next line make the matching based on rigid geometry infeasible.
We state our problem as follows: \textit{given a pair of document images, compute a 
similarity score by detecting patterns of text re-usages between documents 
irrespective of the minor variations in word morphology, word ordering, layout and paraphrasing of 
the content.} Our matching scheme is broadly split into two stages. The first
stage involves segmentation of document into multiple possible word bounding boxes
while the later stage computes a structured document similarity score which obeys
loose word ordering and its content.

%In our matching scheme, we use the observations that (i) our
%images are document images where text is prominent and it is quite valid
%to have the same word repeating in the documents, (ii) There is some amount 
%of intra class variation even within a single document and the best match need 
%not be the true match, (iii) Our interest is finding the near duplicates and the
%documents with very high variation (only few words match) need not be the focus
%in designing the score, (iv) Since the notion of plagiarism or similarity is
%subjective, e.g., how severe paraphrasing can be allowed, the matching scheme should
%allow one to balance the performance with minimal effort or adjustment. 

%\begin{SCfigure}[][t]
%\centering
%\includegraphics[height=3.4cm,width=6cm]{Images/DocSim_Seg.pdf}
%%\fbox{\includegraphics[height=3cm]{Images/HW1k_Seg.pdf}}
%\caption{One of the possible segmentation proposals on a sample document page from DocSim dataset 
%obtained using the proposed segmentation algorithm. Our method is able to arrive at reasonable 
%word segmentation irrespective of challenges such as irregular skew and variable sizing of word 
%bounding boxes}
%\label{fig:segGraph}
%\end{SCfigure}

\subsection{Document segmentation}
\label{subsec:docSeg}
A document image contains structured objects. The objects here are the words and structure 
is the order in which words are presented. 
Segmentation of a handwritten document image into constituent words is a challenging task 
mostly because of the unconstrained nature of documents such as variable placements of page elements, e.g., 
figures, graphs, equations, presence of non-uniform skewed lines, and irregular kerning. 
Most of the methods such as~\cite{GatosSL11,StamatopoulosGLPA13} are bottom-up 
approaches with tunable parameters to arrive at a unique segmentation of words and lines. 
Considering the complexity of handwritten documents, we argue that a reasonably practical 
system, should work with multiple possible lines and word segmentation proposals with 
a high recall, and allow the later modules to deal with combining the results. We use a 
simple multi-stage bottom-up approach similar to~\cite{LouloudisGPH09} by forming three sets 
of connected components (CCs) based on their average sizes. 
%The smaller ones ($s_1$) are assumed 
%to be punctuation, medium ones ($s_2$) the actual characters, and the large ones ($s_3$) 
%could be the components where there is high probability of merges between the lines. 
The small ($s_1$), medium ($s_2$) and large ($s_3$) connected components are assumed to be punctuation, 
actual characters and high probable line merge respectively.
We associate 
each component in $s_2$ with its adjacent component if the cost given in (\ref{eq:ccConnect}), 
is above a certain threshold.
%Fig.~\ref{fig:segGraph} shows a sample page from 
%our collection of classroom assignments along with its proposed segmentation proposals. 

\begin{equation}
Cost(i,j) = OL(i,j) + D(i,j) + \theta(i,j).
\label{eq:ccConnect}
\end{equation}
Here $i,j$ are two components, $OL$ is the amount of overlap in $y$-axis given by intersection over 
union, $D$ is the normalized distance between the centroids of the $i^{th}$ and $j^{th}$ component,
and $\theta(i,j)$ gives the angle between the centroids of the components. After the initial assignment,
we now associate the $s_3$ 
components by checking whether these components intersects in the path of detected lines. In such a case, 
we slice the component horizontally and join it to the top and the bottom line respectively. Finally the components present 
in $s_1$ are associated with nearest detected lines. Given the bounding boxes of a set of CCs and its line associations, 
we analyse the inter CC spacing and derive multiple thresholds to group them into words. This results in 
multiple word bounding box hypotheses with a high recall. Minor reduction in the precision at this stage is taken care 
by our matching scheme. 
%We provide a pseudo code of the proposed segmentation scheme in the supplementary material of this paper.

\subsection{SWM matching}
We first define a similarity score between a pair of documents as the sum of word matches (\textsc{swm}). We use $l_2$ 
normalized {\sc cnn} descriptors of the corresponding 
words images $w_k$ and $w_l$ and compute the $l_2$ distance $d_{kl}$. 
We define the document similarity as the symmetric distance between the best word matches 
across the documents as follows:
\begin{equation}
{\cal S}_N({\cal D}_i, {\cal D}_j)= \frac{1}{|{\cal D}_i|+|{\cal D}_j|} \left ( \sum_{w_k\in {\cal D}_i}
\min_{w_l\in {\cal D}_j} d_{kl} + \sum_{w_l\in {\cal D}_j} \min_{w_k\in
{\cal D}_i} d_{lk} \right ). 
\label{eq:naiveMatch}
\end{equation}
This is a normalized symmetric distance where $|{\cal D}_i|$ is the number of words in the 
document ${\cal D}_i$. In order to reduce the exhaustive matches, we use an approximate nearest 
neighbor search using KD trees.

\begin{figure}[t]
\centering
\includegraphics[width=9cm]{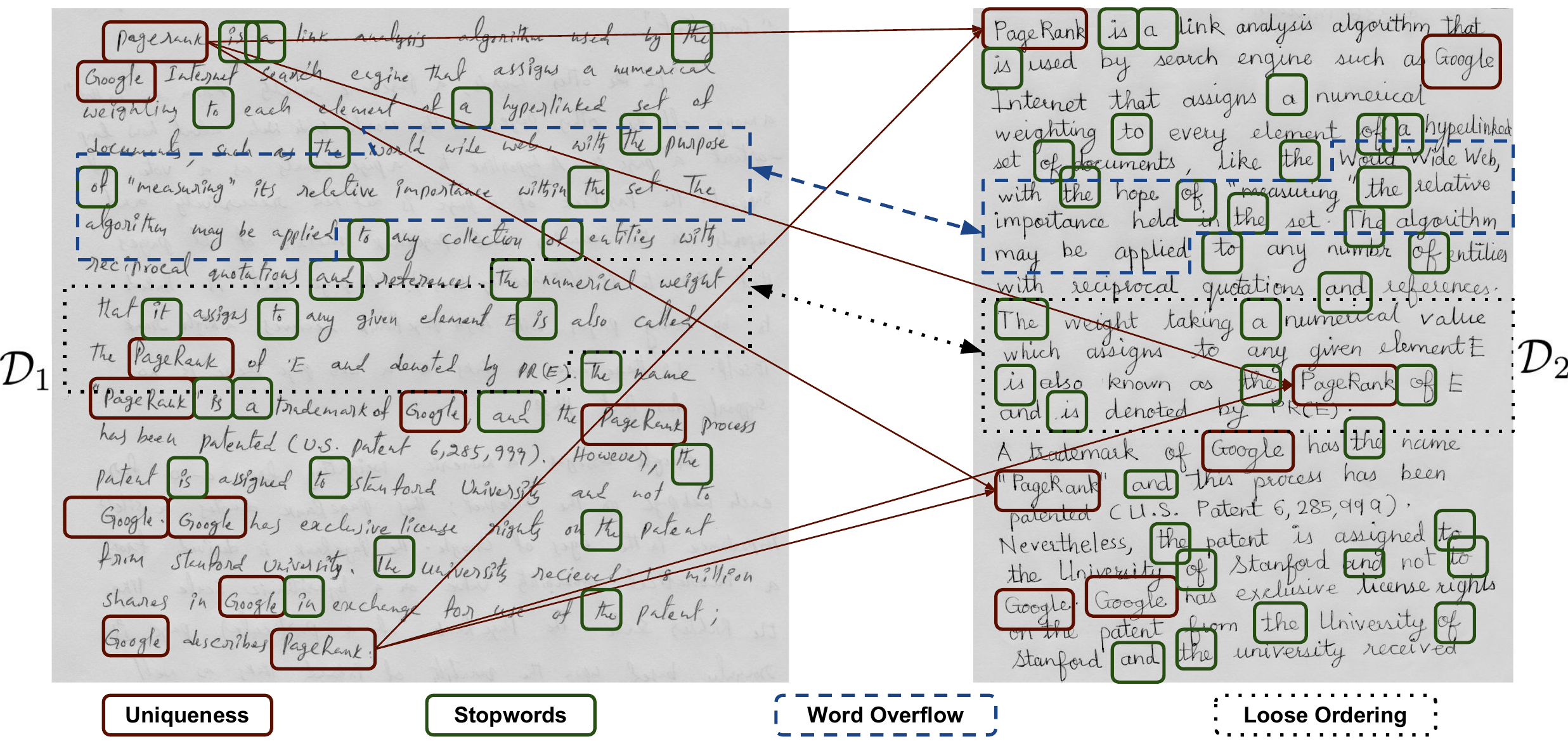}
\caption{A few major challenges of the matching process between a pair of documents ${\cal D}_1$ and ${\cal D}_2$. 
(i) Finding a unique match of each potential word, (ii) removal of stopwords, (iii) invariance to word
overflow problems, and (iv) exploiting the loose ordering of words in matching}
\label{fig:visMatch}
\end{figure}

\subsection{MODS matching}
\label{subsec:dma}
%Fig.~\ref{fig:visMatch} illustrates the problem of document matching  
%and highlights the challenges one needs to resolve for designing an efficient 
%similarity score. 
The problem of document matching and devising a scheme to compute similarity score is a challenging task. 
This problem along with the challenges is illustrated in Fig.~\ref{fig:visMatch}.
We address these problems along with their solution at two levels: (i) individual word matches,
and (ii) bringing locality constraints.

\noindent \textbf{Word matches.} (i) Alternations: In general, the pair of documents of interest need not have
the same content and hence, not all words need to have a correspondence in the second image. 
We enforce this with a simple threshold $\gamma$ on the distance used for matching. 
(ii) Stopwords: The presence of stopwords in documents acts as a noise which corrupts the 
matching process for any IR system due their high frequency. In Fig.~\ref{fig:visMatch} we show 
some of these words in dark green boxes. We observed that the trained HWNet is 
reasonably robust in classifying stopwords due to their limited number and increased presence in training data.
Therefore, we could take the softmax scores (probabilities) from last layer of HWNet and 
classify a word image as a stopword if the scores of one of stopword classes is above a certain threshold.

\noindent \textbf{Locality constraints.} The following three major challenges are addressed using 
locality constraints in the matching process. We first list out the challenges and later propose the solution
given by \textsc{mods}. (i) Uniqueness: Though a word in the first image can match 
with multiple images in the second image, we are interested in a unique match. In Fig.~\ref{fig:visMatch} 
the highlighted words in dark red such as ``Google'' and ``PageRank'' occur at multiple places in both 
documents but the valid matches needs to be unique that obeys the given locality. 
(ii) Word overflow: As we deal with documents of unconstrained nature, similar sentences across different
documents can span variable number of lines, a property of an individual writing style. In terms of 
geometry of position of words this results in a major shift of words (from right extreme 
to the left extreme). One such pair of occurrence is shown in Fig.~\ref{fig:visMatch} as blue colored 
dashed region. We refer to this problem as word overflow.
(iii) Loose ordering: Paraphrasing of the words as shown in the Fig.~\ref{fig:visMatch} as black dashed rectangle,
is a common technique to conceal the act of copying where one changes the order of the words keeping the 
semantics intact. 

%To impose the mentioned constraints into the matching process, 
We observe that the most informative matching words are the ones which preserve the consistency within a locality. We enforce 
locality constraints by splitting the document into multiple overlapping rectangular
regions. The idea is to find out the best matching pairs of regions within two documents
and associate them with individual word matches. For finding the cost of associating two rectangular
regions, we formulate the problem as a weighted bipartite graph matching where the weights are the cosine 
distances of word images in feature space. We use the popular Hungarian algorithm to compute the cost of 
word assignments, which leads to a one to one mapping of word images between a pair of regions.
The score computed between a pair of rectangular regions denoted as $p$ and $q$ from documents ${\cal D}_i$
and ${\cal D}_j$ respectively as given by: 
\begin{equation}
\centering
Score(p) = \max_{q \in R({\cal D}_j)} \left(\frac{\sum_{(k,l)\in Matches(p,q)}(1-d_{kl})}{max(|p|,|q|)}\right), \forall p \in R({\cal D}_i),
\label{eq:regionScore}
\end{equation}
where, $R({\cal D}_j)$ denotes 
the set of all rectangular regions in a document image. The function $Matches(p,q)$ returns the 
assignments given by the Hungarian algorithm. Finally, the normalized \textsc{mods} score for a pair of 
documents is defined as follows:
\begin{equation}
\centering
{\cal S}_M({\cal D}_i,{\cal D}_j) = \frac{\sum_{p \in R({\cal D}_i)} Score(p)}{max(|{\cal D}_i|,|{\cal D}_j|)}.
\label{eq:modsScore}
\end{equation}
We provide the pseudo-code of the proposed matching framework 
in the supplementary material.

%This is taken care by limiting the ordering as if $A$ matches with $B$, then a match from the neighbourhood of $A$ 
%will also be in the neighbourhood of $B$. 
%It also draws contrasts from the conventional matching problems encountered in vision. 
%Based on our observations, we impose the following constraints on the matching process.
%This is also one of the reasons why we cannot use the
%traditional vision based matching schemes where the rigidity of geometry is necessary.
%This is addressed by treating the matching of the words in a  {\em near} linear manner. 

%\renewcommand{\arraystretch}{1.5pt}
\begin{table*}[t]
\setlength\heavyrulewidth{0.15em}
\setlength\lightrulewidth{0.10em}
\centering
%\begin{minipage}[t]{\textwidth}
    \caption{Quantitative results on word spotting using the proposed \textsc{cnn} features along with comparisons 
    with various existing hand designed features on \textsc{iam} and \textsc{gw} dataset}
    \label{tab:wordSpot}
    \begin{tabularx}{0.75\textwidth}{@{} X c c c c c c c @{}}
    \toprule
    Dataset & \textsc{dtw}\cite{AlmazanPAMI14} & \textsc{sc-hmm}\cite{Rodriguez-SerranoP12} & \textsc{fv}\cite{AlmazanPAMI14} & \textsc{ex-svm}\cite{AlmazanPR14} & \textsc{kcca}\cite{AlmazanICCV13} & \textsc{kcsr}\cite{AlmazanPAMI14} & Ours \\ \midrule
    {\sc gw} & 0.6063 & 0.5300 & 0.6272 & 0.5913 & 0.8563 & 0.9290 & \textbf{0.9484} \\
    {\sc iam} & 0.1230 & - & 0.1566 & - & 0.5478 & 0.5573 & \textbf{0.8061} \\
    \bottomrule
    \end{tabularx}
%\end{minipage}
\end{table*}

\section{Experiments}
\label{sec:exp}
In this section, we empirically evaluate the proposed {\sc cnn} representation for the task of word spotting
on standard datasets. We validate the effectiveness of these features on newer tasks such as retrieving semantically
similar words, searching in keywords from instructional videos and finally demonstrate the performance 
of the {\sc mods} algorithm for finding similarity between documents on annotated datasets created for this purpose. 

\subsection{Word-spotting}
\label{subsec:wordSpot}
\begin{figure}[t]
\centering
\includegraphics[height=4cm]{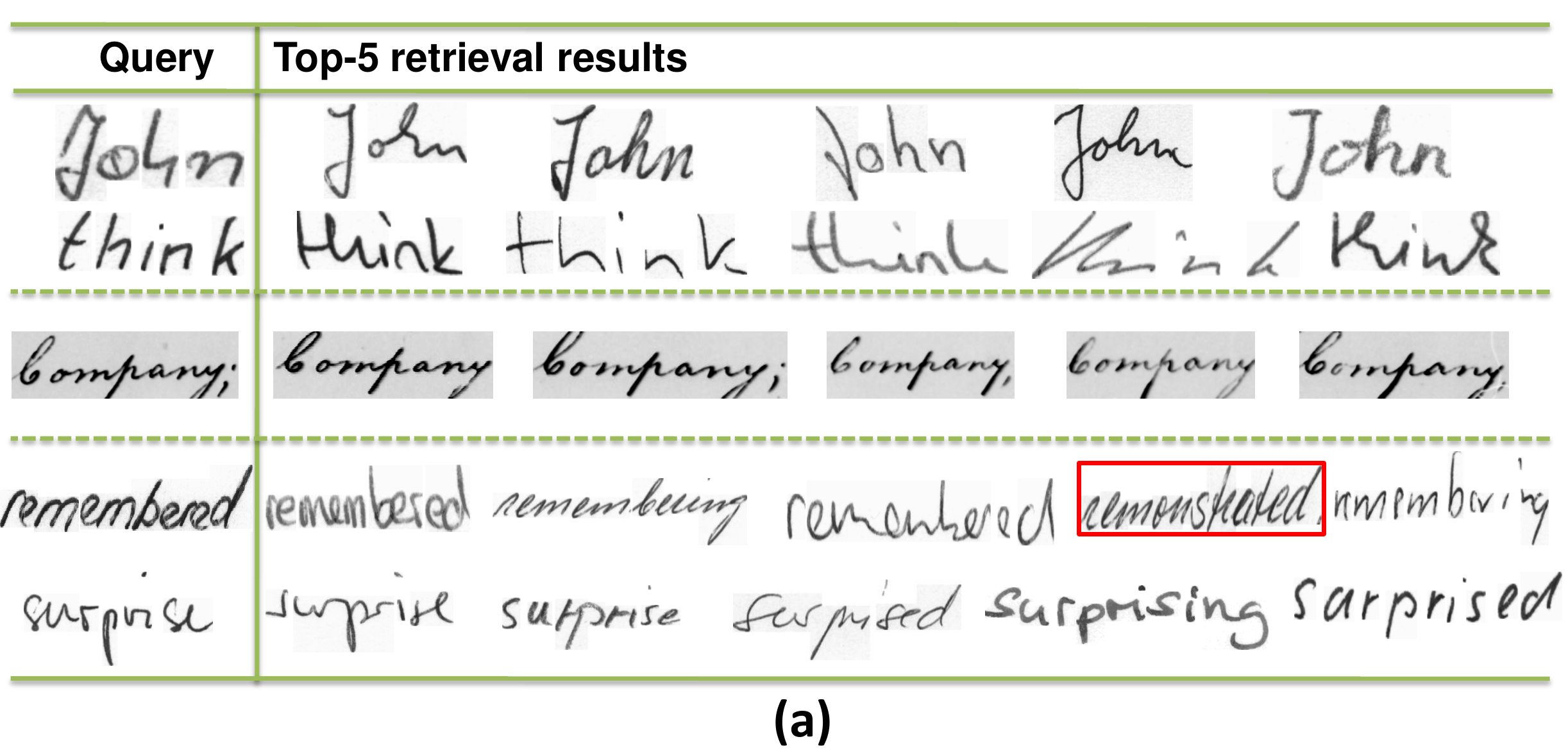}
%\quad
\includegraphics[height=4cm]{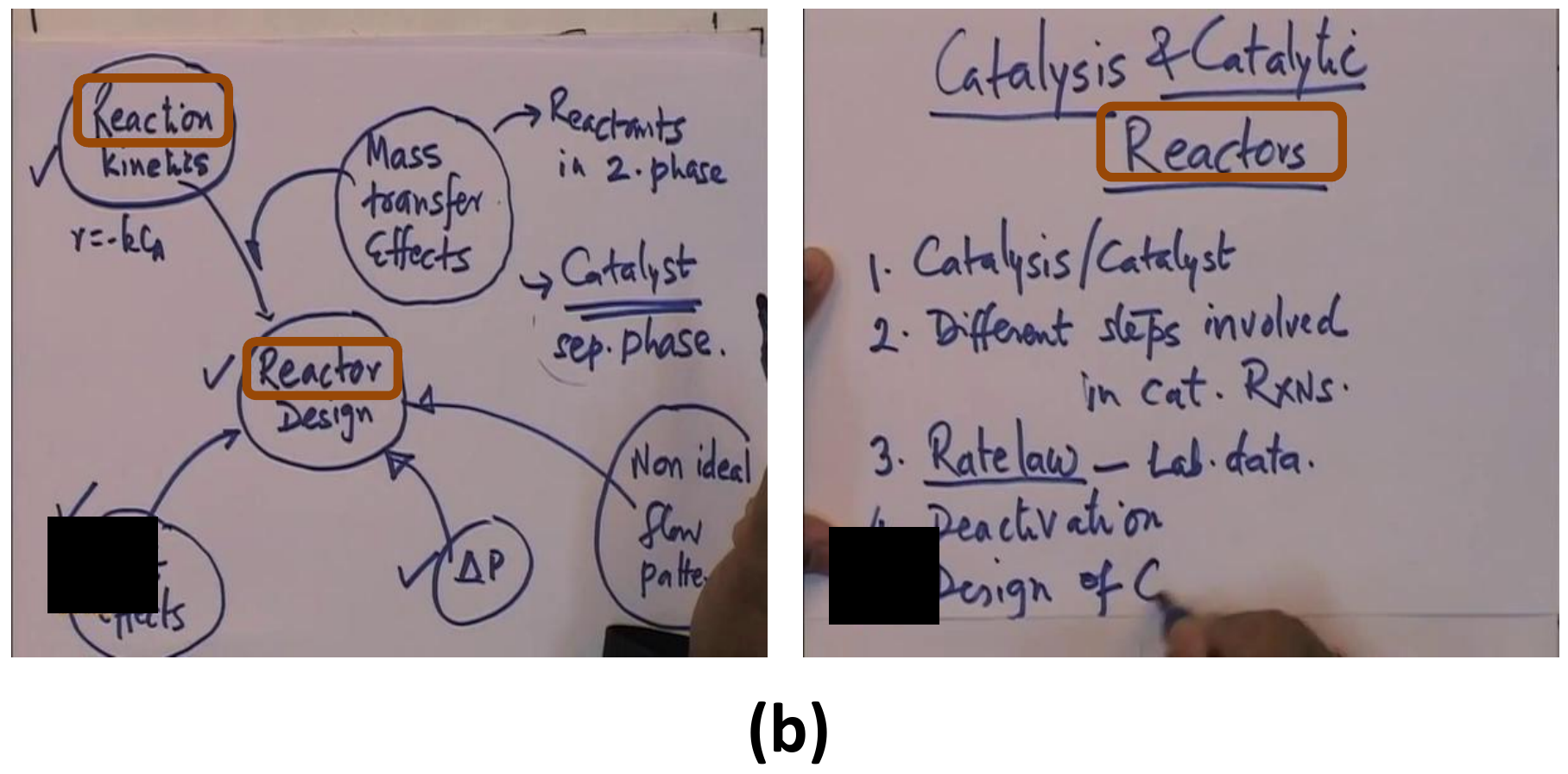}
\caption{Qualitative results: (a) Query-by-example results for the task of word spotting results on \textsc{iam} and 
\textsc{gw} dataset. The bottom two rows shows results from normalized feature representation where one can observe 
we are also able to retrieve words with related meanings. (b) Query-by-text results on searching with ``reactor''
on an instructional video. The top two results are shown along with the spotted words which are highlighted in the frame}
%\vspace{-0.4cm}
\label{fig:qualWordSpot}
\end{figure}

We perform word spotting in a \textit{query-by-example} setting. 
%where given a query word image, the objective is to find the most similar word images in a ranked order. 
We use \textsc{iam}~\cite{Marti02} and George Washington~\cite{Fischer12} (\textsc{gw}) dataset, popularly used in handwritten word spotting and recognition tasks. 
In case of the \textsc{iam} dataset, we use the standard partition for training, testing, and validation provided along 
with the corpus. For \textsc{gw} dataset, we use a random set of $75\%$ for training and validation, and 
the remaining $25\%$ for testing. Each word image except the stop words in the test corpus is taken as the query to be 
ranked across all other images from the test corpus including stop-words acting as distractions. The performance 
is measured using the standard evaluation measure namely, mean Average Precision (mAP). HWNet architecture  
is fine-tuned using the respective standard training set for each test scenario. Table~\ref{tab:wordSpot} 
compares the proposed features from state-of-the-art methods on these datasets. The results are evaluated in
a case-insensitive manner as used in previous works~\cite{AlmazanPR14,AlmazanPAMI14}.
%The performance of methods such as \textsc{dtw}~\cite{Rodriguez-SerranoP12} and \textsc{hmm} are clearly 
%inferior to the recent methods. 
%Handcrafted features such as Fisher vector (\textsc{fv}) did not perform well on multi-writer handwritten images. 
The proposed \textsc{cnn} features clearly surpasses the current state-of-the-art method~\cite{AlmazanPAMI14}
on \textsc{iam} and \textsc{gw}, reducing the error rates by $\sim55\%$ and $\sim37\%$ respectively. 
This demonstrates the invariance of features for both multi-writer scenario (\textsc{iam}) and historical documents 
(\textsc{gw}). Some of the qualitative results are shown in the top three rows of Fig.~\ref{fig:qualWordSpot}(a). One can
observe the variability of each retrieved result which demonstrates the robustness the proposed features.

\begin{figure}[t]
\centering
\includegraphics[scale=0.18]{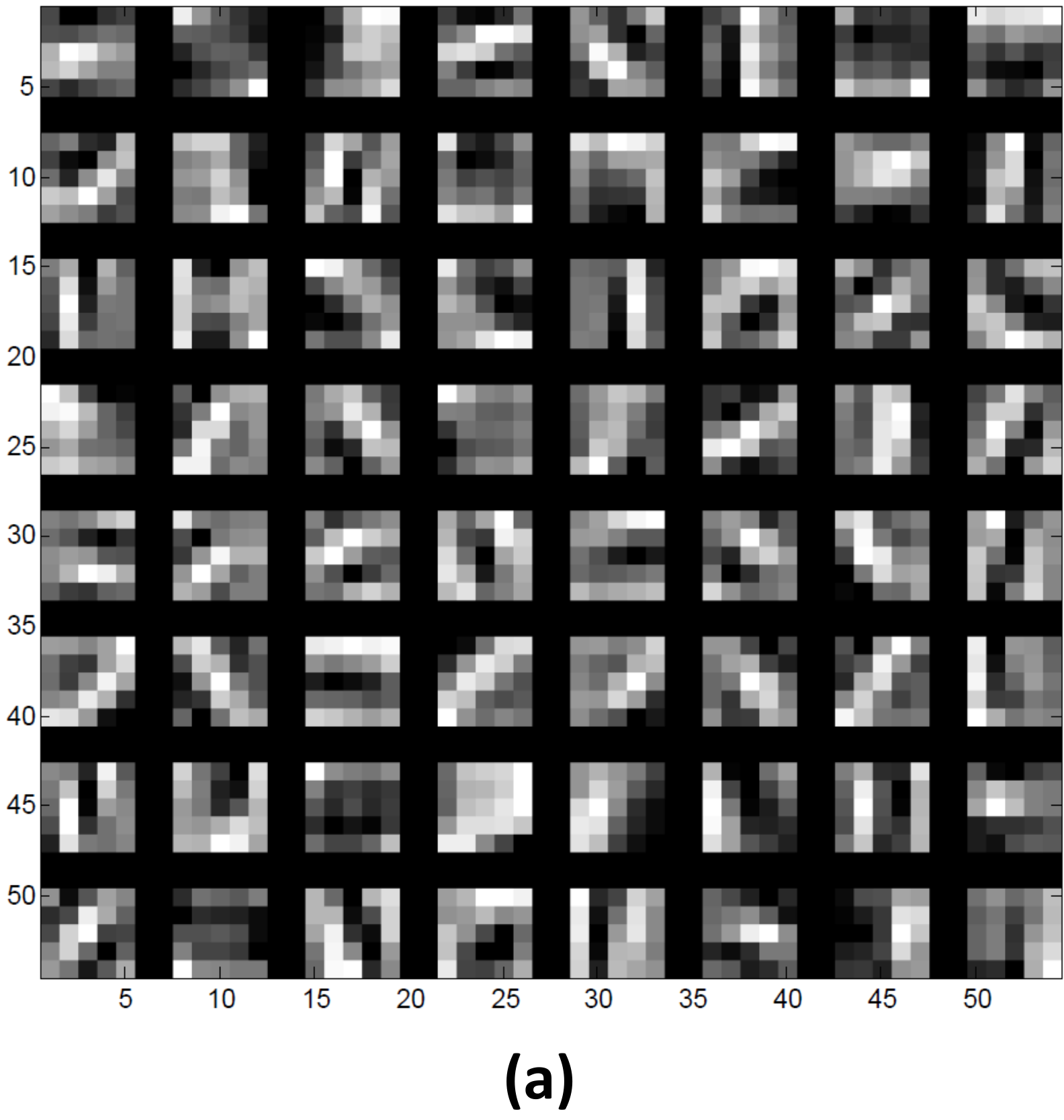}
\includegraphics[scale=0.2]{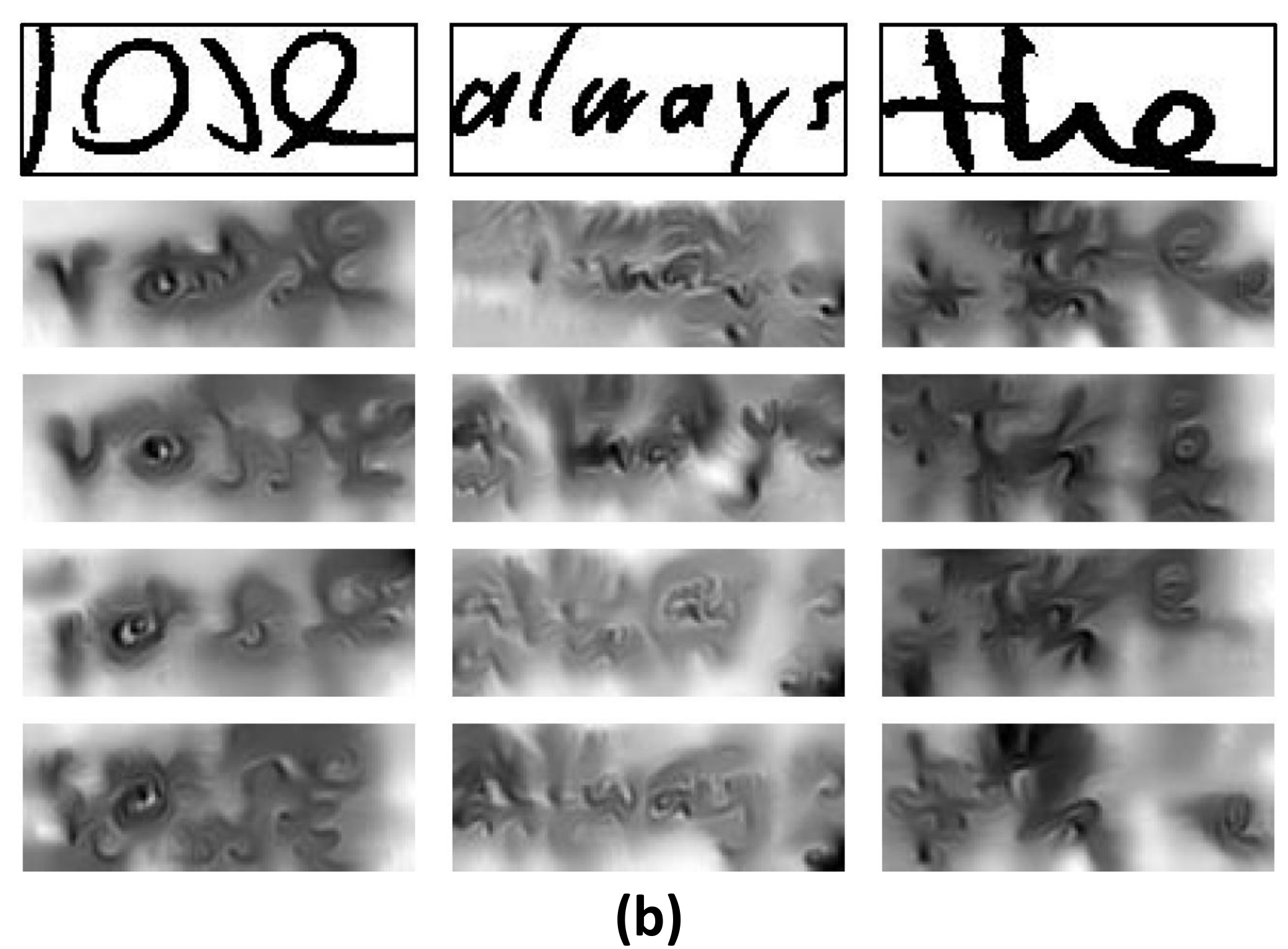}
\caption{Visualization: (a) The weights of first layer of HWNet. (b) Four possible reconstructions~\cite{MahendranV15} 
of three sample word images shown in columns. These are re-constructed from the representation of final layer of HWNet}
\label{fig:visLayers}
\end{figure}
%\begin{table}[t]
%\setlength\heavyrulewidth{0.15em}
%\setlength\lightrulewidth{0.10em}
%\centering
%    \begin{minipage}[t]{0.52\linewidth}
%        \caption{Comparison of performance improvement by fine-tuning the existing architectures
%        and the proposed architecture on~\textsc{iam} dataset}
%        \label{tab:cnnAdapt}
%        \renewcommand\arraystretch{1.27}
%        \begin{tabularx}{\textwidth}{@{} X c c c @{}}
%        \toprule
%        Architecture & {\sc orig} & {\sc orig}+{\sc f} & {\sc hw-synth}+{\sc f}\\ \midrule
%        AlexNet.~\cite{KrizhevskySH12}  & 0.2997 & 0.4468 & 0.5786 \\
%        JSVZNet.~\cite{JaderbergSVZ14}  & 0.3746  & 0.4822 & 0.5022\\
%        HWNet                           & 0.5386 & * & \textbf{0.8061}\\
%        \toprule
%        \end{tabularx}
%    \end{minipage}
%    \quad
%    \begin{minipage}[t]{0.43\linewidth}
%        \caption{Word spotting results using normalized features and its comparisons with exact features.}
%        \label{tab:wordForm}
%        \renewcommand\arraystretch{}
%        \begin{tabularx}{\textwidth}{@{} X c c c @{}}
%        \toprule
%        \multirow{2}{*}{Evaluation} & \multicolumn{2}{c}{Features} \\
%        & \textsc{cnn} & \textsc{cnn}$_{Norm}$\\ \midrule
%        Exact & \textbf{0.8061} & 0.7955 \\
%        Inexact & 0.7170 & \textbf{0.7443} \\
%        \toprule
%        \end{tabularx}
%    \end{minipage}
%\end{table}

\subsubsection{Visualizations.}
\label{subsubsec:visualize}
Fig.~\ref{fig:visLayers} shows the visualization of the trained HWNet architecture using popular schemes demonstrated 
in~\cite{MahendranV15,KrizhevskySH12}. Fig.~\ref{fig:visLayers}(a) visualizes the weights of the first layer which bears a resemblance
to Gabor filters and detects edges in different orientations. Fig.~\ref{fig:visLayers}(b) demonstrates the visualization
from a recent method~\cite{MahendranV15} which inverts the {\sc cnn} encoding back to image space and arrives at possibles 
images which have high degree of probability for that encoding. This gives a better intuition of the learned layers and 
helps in understanding the invariances of the network. Here, we show the query images on the first row and its reconstruction in
the following rows. One can observe that in almost all reconstructions there are multiple translated copies of the characters
present in the word image along with some degree of orientations. Similarly, we can see the network is invariant to the first letter 
being in capital case (see Label: ``the'' at Col:3, Row:4) which was part of the training process. The reconstruction of the first 
image (see Label: ``rose'' at Col:1, Row:1) shows that possible reconstruction images includes Label: ``rose'' (Col:1, Row:2) and 
``jose'' (Col:1, Row:3) since there is an ambiguity in the query image. In the supplementary material, we also study HWNet architecture 
by comparing it with other popular deep learning models such as AlexNet~\cite{KrizhevskySH12} and scene text recognition 
model of~\cite{JaderbergSVZ14}.

\subsection{Enhancements and applications}
\label{subsec:novelTasks}
We now analyse the performance of the normalized features for retrieving semantically similar words which has not been
yet attempted in handwritten domain and plays an important role in matching similar documents. We also demonstrate 
an application of \textsc{mods} framework in a collection of instructional videos by retrieving relevant frames 
corresponding to user queries.
\begin{table}
\centering
\setlength{\tabcolsep}{8pt}
    \caption{Word spotting results using normalized features and its comparisons with exact features.}
    \begin{tabular}{|c|c|c|}
    \hline
    %\multirow{2}{*}{Evaluation} & \multicolumn{2}{c|}{Features} \\\cline{2-3}
     Evaluation &  \textsc{cnn} & \textsc{cnn}$_{Norm}$\\\hline
    Exact & \textbf{0.8061} & 0.7955 \\\hline
    Inexact & 0.7170 & \textbf{0.7443} \\\hline
    \end{tabular}
    \label{tab:wordForm}
\end{table}

\subsubsection{Normalized word spotting.}
\label{subsubsec:normWordSpot}
Table~\ref{tab:wordForm} shows the quantitative results of the normalized (\textsc{cnn}$_{Norm}$) features which are 
invariant to common word inflectional endings and thereby learn features for stem or the root part of the word image. 
For this experiment, we update the evaluation scheme (ref. as inexact) to include not only similar word images but 
also the word images having common stem. We use Porter stemmer~\cite{porter80} for 
calculating the stem of a word. Table~\ref{tab:wordForm} also compares the performance of \textsc{cnn}
features used in Sec.~\ref{subsec:wordSpot}
and validate it over inexact evaluation. Here we obtain a reduced mAP of $0.7170$ whereas using the normalized features,
we improve the mAP to $0.7443$. We also observe that using normalized features for exact evaluation results in a comparable
performance ($0.7955$) which motivates us to use them in document similarity problems. In Fig.~\ref{fig:qualWordSpot}(a), the 
bottom two rows shows qualitative results using these normalized features. The retrieval results for query ``surprise'' 
contains the word ``surprised'', ``surprising'' along with the keyword ``surprise''.

\subsubsection{Searching in instructional videos.}
\label{subsubsec:videoSearch}
To demonstrate the effectiveness and generalization ability of the proposed \textsc{cnn} features we performed an interesting
task of searching inside instructional videos where the tutor write handwritten text to aid students in the class. We conducted
the experiment in a query-by-text scenario where the query text is synthesized into a word image using one of the fonts used in the \textsc{hw-synth}
dataset. We took 5 popular online course videos on different topics from YouTube and extracted keyframes from them. 
%In the current work,
%we took the videos where the content is written on paper or white board and the camera is capturing the top view of the writing medium.
We obtained multiple segmentation output from the proposed segmentation method. 

For evaluation, we handpicked 20 important queries and labeled
the frames containing them. We obtained a frame level mAP of $0.9369$ on this task. Fig.~\ref{fig:qualWordSpot}(b) shows the top-2 matching frames
for the query ``reactor'' along with the spotted words. One can observe that along with retrieving exact matches, 
we also retrieve similar keywords such as ``Reactors'', and ``Reaction''.

\subsection{DocSim dataset and evaluations}
\label{subsec:assignEval}
%DocSim dataset is prepared as a part of a virtual assignment given to a set of students. 
%The content over which the document was prepared was controlled in-order to quantify the 
%amount of similarity it possessed. 
We start with the textual corpus presented in~\cite{CloughLRE11} 
for plagiarism detection.
The corpus contains plagiarized short answers to five unique questions given to 19 participants. 
Hence the corpus contains around 100 documents of which 95 were created in a controlled setting 
while five were the original answers (source document) which were given to participants
to refer to and copy. There are four types or degree of plagiarism introduced in this 
collection: (i) \textit{near copy}, where the content is an exact copy from different parts from 
the source; (ii) \textit{light revision}, where the content is taken from source but with slight 
revisions such as replacing words with synonyms, (iii) \textit{heavy revision}, which includes heavy 
modification such as paraphrasing, combining or splitting sentences and changing the order; and (iv) 
\textit{non-plagiarized}, where the content is prepared independently on the same topic. For the 
task of generating handwritten document images, we included a total of 24 students and asked them 
to write on plain white sheets of paper. For each document we use a separate student to avoid any biases in 
writing styles. To keep the content close to its natural form, we did not mention any requirements 
on spacing between words, and lines, and did not put any constraints on the formatting of text in the form 
of line breaks and paragraphs. In case of mistakes, the written word was striked out and writing 
was continued. 
%Each answer written can span multiple A4 sheets and was written in a back-to-back. 
%The top two images in Fig.~\ref{fig:visDataset} shows a sample source document (part of the document) along 
%with the plagiarized version written by a different student and has undergone light revisions in the content.

%In text domain, under the area of plagiarism detection, 
%there are some standard datasets available such as~\cite{PotthastCOLING10,CloughLRE11}. 
%hence we used it to re-create the corresponding handwritten documents. 

%Each participant is given 5 different questions to answer along with 
%the \textit{source} documents (answer to each questions) which should be used for copying. 

\begin{table*}[t]
\centering
\setlength{\tabcolsep}{8pt}
    \caption{Quantitative evaluation of various matching schemes on DocSim dataset. We compare the performance 
    of proposed \textsc{mods} framework using \textsc{cnn} features over baseline methods such as \textsc{nn},
    \textsc{bow}, and embedded attributes proposed in~\cite{AlmazanPAMI14}}

    \begin{tabular}{|c||c c||c c||c c|c}
    \hline
    \textbf{Method} & \textsc{nn} & \textsc{bow} & \textsc{swm} & {\sc mods} & \textsc{swm} & {\sc mods}\\\hline
    \textbf{Feature} & Profile & \textsc{sift} & \multicolumn{2}{c||}{\textsc{kcsr}~\cite{AlmazanPAMI14}} & \multicolumn{2}{c|}{\textsc{cnn}} \\\hline
    \textbf{$nDCG@99$} & 0.5856 & 0.6128 & 0.7968 & 0.8444 & 0.8569 & \textbf{0.8993} \\\hline
    \textbf{AUC} & 0.5377 & 0.4516 & 0.8231 & 0.8302 & 0.9465 & \textbf{0.9720} \\\hline
    \end{tabular}
    \label{tab:qResults}
\end{table*}

\begin{figure}[t]
\centering
%\begin{minipage}[b]{0.45\textwidth}
    \fbox{\includegraphics[width=8cm]{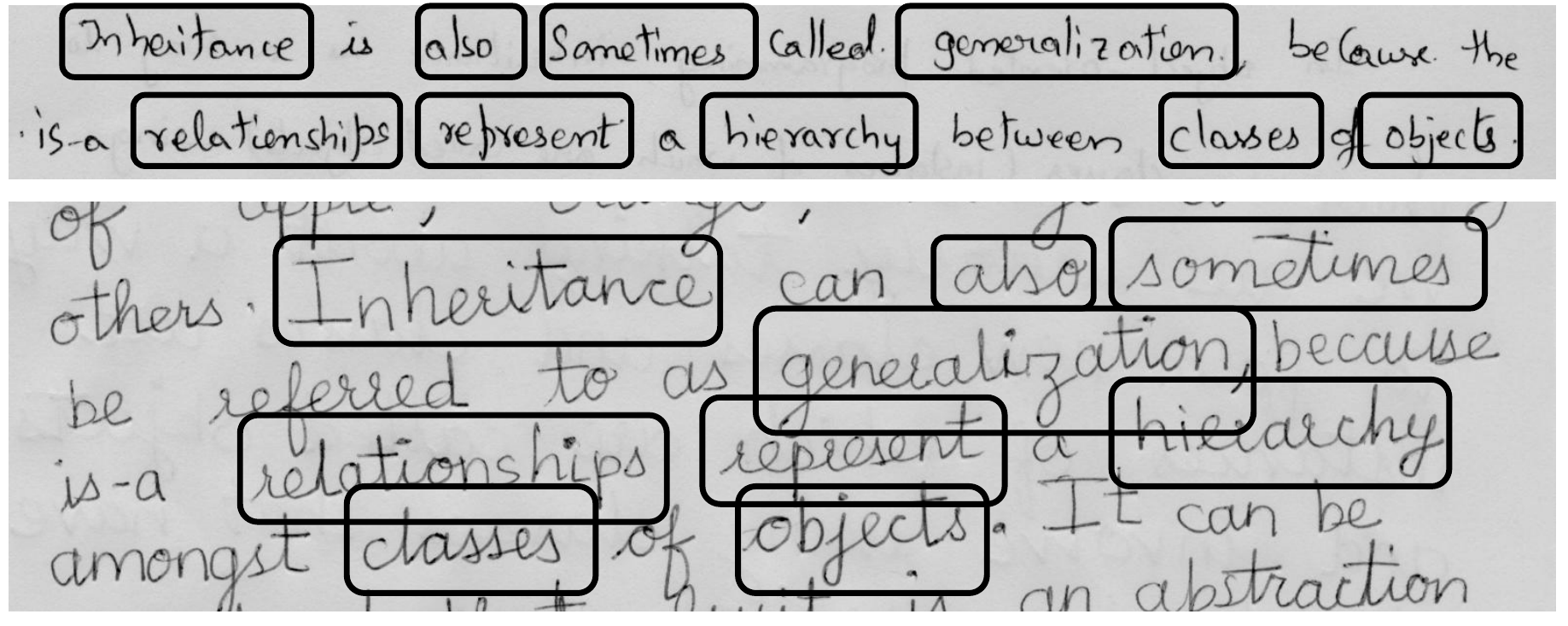}}
    
%\end{minipage}
%\begin{minipage}[b]{0.5\textwidth}
    \fbox{\includegraphics[width=8cm]{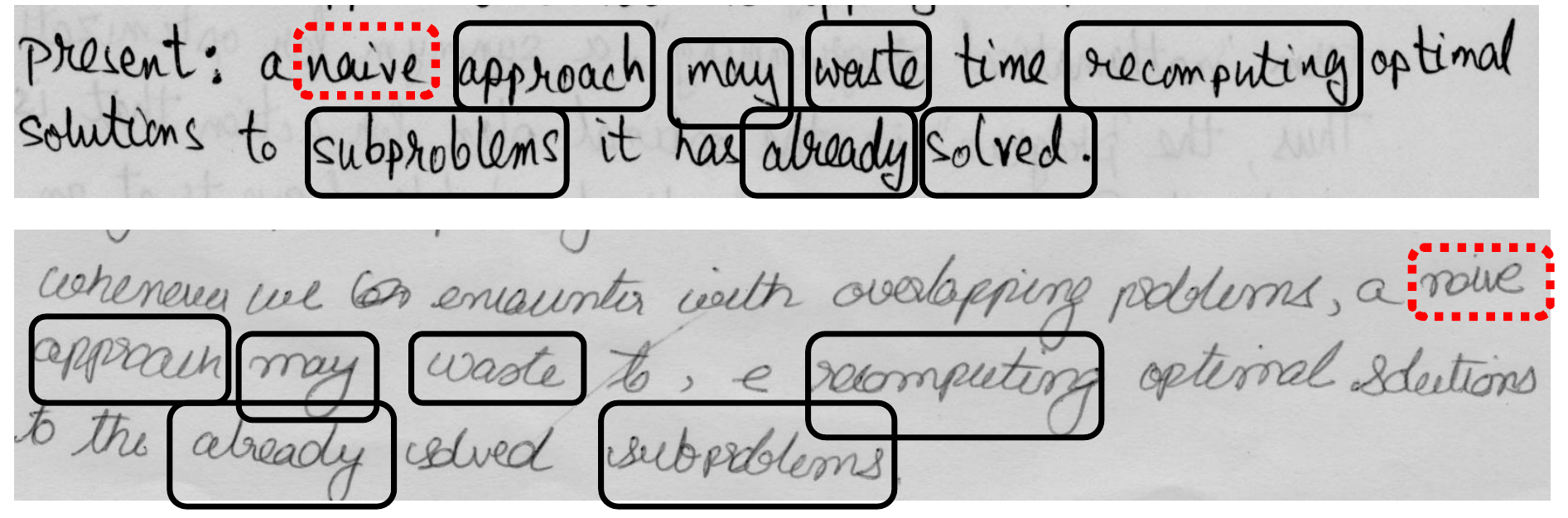}}
%\end{minipage}
\caption{Qualitative results of the \textsc{mods} matching algorithm from DocSim dataset. Here we show two sample
matching pairs images. The top region is taken from source and bottom one is plagiarized. The highlighted
words in rectangle have been correctly matched along with few words which remain undetected.}
%In both cases the proposed method is able to detect valid pairs inspite of style and layout variations
%\vspace{-0.5cm}
\label{fig:qualResults}
\end{figure}
\noindent \textbf{Evaluation methodology.} To evaluate the performance,
we took all source-candidate document pairs and computed their similarity scores. Here we only verify  
whether the document is similar (plagiarized) or not while discarding the amount of plagiarism. The performance is 
measured using area under the ROC 
curve (AUC) by sorting the scores of all pairs. 

In another experiment, we compute graded similarity 
measure in accordance to each source document posed as a query which expects the ranking according to
the degree of copying. Here we use normalized discounted cumulative gain ($nDCG$), a measure 
used frequently in information retrieval when grading is needed. 
Here the query is presented as the \textit{source} document and the target documents 
are all documents present in the corpus. The discounted cumulative gain ($DCG$) at position $p$ is given 
as $DCG_p=\sum_{i=1}^{p} (2^{rel_i} - 1) / (log_2 (i+1))$ where $rel_i$ is the ground truth relevance 
for the document at rank $i$. In our case, the relevance measures are represented as: 3 - \textit{near copy}, 
2 - \textit{light revision}, 1 - \textit{heavy revision}, and 0 - \textit{not copied}. The normalized 
measure $nDCG$ is defined as $DCG_p/IDCG_p$, where $IDCG$ is the $DCG$ measure for ideal ranking. $nDCG$ 
values scale between $0.0-1.0$ with $1.0$ for ideal ranking.
\vspace{0.3cm}

\noindent \textbf{Results.} We now establish two baselines for comparison. 
Our first approach uses a classical visual bag of words ({\sc bow}) 
approach computed at the interest points. The {\sc bow} representation has been 
successfully used in many image retrieval tasks including the document images~\cite{Yalniz12,Shekhar13}. We
use \textsc{sift} descriptors, quantized using \textsc{llc} and represented using a spatial pyramid of size $1\times3$.
Our second baseline ({\sc nn}) uses the classical word spotting scheme based on profile features similar to~\cite{RathM07}. 
While the first one is scalable for large datasets, the second one is not really appropriate due to the time complexity of classical {\sc dtw}. 
In both these methods, the best match is identified as the document which has most number of word/patch matches.
Table~\ref{tab:qResults} reports the quantitative evaluation for various matching schemes along with the baselines.
The proposed {\sc mods} framework along with {\sc cnn} features performs better in both evaluation measures
consistently. Using \textsc{swm} word matching scheme over the proposed \textsc{cnn} features, we achieve an $nDCG$ 
score of $0.8569$ and AUC of $0.9465$. This is further improved in the \textsc{mods}, which incorporates loose 
ordering and is invariant to word overflow problems. Note that in both cases (\textsc{swm} and \textsc{mods}), the stopwords
are removed as preprocessing. We also evaluate our framework with the state-of-the-art features proposed in~\cite{AlmazanPAMI14}
and observe a similar trend which validates the effectiveness of {\sc mods}. Fig.~\ref{fig:qualResults} shows some
qualitative results of matching pairs from DocSim dataset.
%Since matching 
%of the documents based on geometry defined over the interest points, is not a reasonable approach for the 
%{\sc hw} documents, we consider methods based on matching the image segments on words loosely. 
%Fig.~\ref{fig:visDocSim} 
%shows sample matching regions from plagiarized documents found correctly by \textsc{mods}. The matching words in this
%neighbourhood are highlighted. 

%\vspace{-0.3cm}
\subsection{HW-1K dataset and evaluations}
\label{subsec:HW-1KEval}
To validate the performance of the system on an unrestricted collection, we introduce HW-1K dataset 
which is collected from the real assignments of a class as part of an active course. The dataset 
contains nearly 1K handwritten pages from more than 100 students. The content in these documents 
varied from text, figures, plots and mathematical symbols. Most of the documents follow a complex 
layout with misalignment in paragraphs, huge variations in line and word spacing and a high degree of 
skewness over the content. The scanned images also possessed degradation in quality due to loose 
handling by the students which created folds and noise over the paper. The {\sc mods} system 
captures the handwritten documents using a mobile app which scans the physical pages, and uploads 
it to a central server where document images are matched.

\noindent \textbf{Evaluation and results.} We perform a human evaluation 
where we picked a set of 
50 assignment images written by different students, and gathered the top-1 similar document image present in 
the corpus using \textsc{mods}. We then ask five humans evaluators to give a score to the top-1 retrieval on a 
likert scale of $0-3$ where $0$ is ``\emph{very dis-similar}'', $1$ is ``\emph{similar only for few word matches}'', $2$ 
is ``\emph{partially similar}'' and $3$ is ``\emph{totally similar}''. Here, the scale-1 refers to the case where 
the document pair refers to the same topic. Thus there could be individual word matches but the text is not plagiarized. The average 
agreement to the human judgments as evaluated for the top-1 similar document is reported at $2.356$ 
with $3$ as the best score.
%The lower part of Fig.~\ref{fig:qualResults} shows one such instance of matching documents 
%detected by \textsc{mods}. Here we observe that despite the style differences in both documents along with the changes in word order, 
%our method successfully matches it.

%\vspace{-0.3cm}
\section{Discussions}
\label{sec:conc}
We propose a method which estimates a measure of similarity for two handwritten
documents. Given a set of digitized handwritten documents, we estimate a ranked list of similar
pairs that can be used for manual validation, as in the case of {\sc moss} and deciding the
amount of plagiarism. Our document similarity score is computed using a {\sc cnn} feature
descriptor at the word level which surpasses the state-of-the-art results for the task of word spotting
in multi-writer scenarios. We believe that with an annotated, larger set of natural handwritten
word images, the performance can be further improved. We plan to use weakly supervised learning 
techniques for this purpose in the future. 

Throughout this work, we characterize the document images with textual
content alone. Many of the document images also have graphics. Our method fails to compare
them reliably. On a qualitative analyses of the failures, we also find that the performance 
of matching mathematical expressions e.g., equations and symbols is inferior to the textual content. 
We believe identifying regions with graphics and applying separate scheme for matching such regions 
can further enhance the performance of our system.

%Also human handwritten documents often append the content with arrows
%(directions) and inserting a set of lines or words at a different location in a page. 
%Our method fails to take care of such directives. 

\bibliographystyle{IEEEtran}
\bibliography{praveen}

% that's all folks
\end{document}